\documentclass[a4paper,english,nofontune]{cccconf}
\usepackage[comma,numbers,square,sort&compress]{natbib}
\usepackage{epstopdf}
\usepackage{amsmath,amssymb,amsfonts}
\usepackage{algorithm}
\usepackage{algpseudocode}
\usepackage{graphicx}
\usepackage{textcomp}
\usepackage{color}
\usepackage{bm}
\usepackage{comment}
\usepackage{makecell}
\usepackage{footnote}
\usepackage{subfigure}
\usepackage{balance}
\usepackage{microtype}

\begin{document}

\title{Composite FORCE Learning of Chaotic Echo State\\Networks for Time-Series Prediction}

\author{Yansong Li\aref{AI},
        Kai Hu\aref{AI},
        Kohei Nakajima\aref{UTO}, and
        Yongping Pan$^{*}$\aref{AI,CSE}
        }


\affiliation[AI]{School of Artificial Intelligence,
        Sun Yat-sen University, Zhuhai 519000, China
        \email{liys33@mail2.sysu.edu.cn;
        hukai8@mail2.sysu.edu.cn; panyongp@mail.sysu.edu.cn}}
 \affiliation[CSE]{School of Computer Science and Engineering, Sun Yat-sen University, Guangzhou 510006, China}
 \affiliation[UTO]{Graduate School of Information Science and Technology, University of Tokyo, Tokyo 113-8656, Japan
         \email{k\_nakajima@mech.t.u-tokyo.ac.jp}}
\maketitle

\begin{abstract}

Echo state network (ESN), a kind of recurrent neural networks, consists of a fixed reservoir in which neurons are connected randomly and recursively and obtains the desired output only by training output connection weights. First-order reduced and controlled error (FORCE) learning is an online supervised training approach that can change the chaotic activity of ESNs into specified activity patterns. This paper proposes a composite FORCE learning method based on recursive least squares to train ESNs whose initial activity is spontaneously chaotic, where a composite learning technique featured by dynamic regressor extension and memory data exploitation is applied to enhance parameter convergence. The proposed method is applied to a benchmark problem about predicting chaotic time series generated by the Mackey-Glass system, and numerical results have shown that it significantly improves learning and prediction performances compared with existing methods.

\end{abstract}

\keywords{Chaotic Neural Network, Recurrent Neural Network, FORCE Learning, Composite Learning, Chaotic System}

\footnotetext{*Corresponding author. This work was supported in part by the Guangdong Pearl River Talent Program of China under Grant No. 2019QN01X154.}

\section{Introduction}
Recurrent neural networks (RNNs) with feedback loops can maintain ongoing activations even in the absence of external inputs \cite{Jaeger2004Harnessing}. In theory, RNNs can learn to mimic any dynamical system with arbitrary accuracy \cite{Funahashi1993Approximation}. An RNN can preserve a nonlinear transformation of its historical inputs in its internal states, thus exhibiting dynamical memory for temporal context information processing \cite{Luko2009Reservoir}. In this sense, RNNs are highly promising and fascinating for nonlinear time-series modeling. Nevertheless, traditional training approaches of RNNs suffer from some critical problems such as gradient information degeneration, long training time, slow convergence, and local minima \cite{Doya1992Bifurcations, Bengio2002Learning, Atiya2000New, Jaeger2002Tutorial}. Traditional training approaches may not even converge when RNNs predict chaotic activities with irregular initial conditions and exponential sensitivity \cite{Abarbanel2008Estimation}.
Reservoir computing (RC) is a more accessible framework for designing and training RNNs \cite{Verstraeten2007RCmethods}. A salient feature of RC is that reservoir neurons are randomly generated and sparsely connected, and thereby only the readout layer needs to be trained. The realization of RC in traditional RNNs results in echo state networks (ESNs) \cite{Jaeger2001echostate}, and liquid state machines can be regarded as a spiking-neuron version of ESNs \cite{Maass2002Realtime}.
External inputs and neuron activations can activate an ESN to obtain rich echo states.
Because each neuron has its own nonlinear activity, the reservoir can map network inputs to a higher-dimensional space, and the readout can produce the output from the reservoir by a linear combination of neuron activations. Connection weights among the reservoir neurons stay fixed during training, and only connection weights between the reservoir and the output unit are trained to compute the desired output \cite{Schrauwen2007overviewRC, Luko2009Reservoir, Luko2012Reservoir, nakajima2021reservoir}. Under this circumstance, the shortcomings of RNN training approaches based on error backpropagation can be largely avoided. ESNs have greatly facilitated the practical applications of RNNs and outperform fully trained RNNs in several applications \cite{wu2021adaptive}.

ESNs can be trained offline \cite{Jaeger2004Harnessing} or online \cite{Jaeger2002Adaptive}. Sussillo et al. \cite{Sussillo2009Generating} proposed an online supervised training method called “first-order reduced and controlled error (FORCE)” learning to take advantage of an initially spontaneously chaotic RNN and a feedback loop between the network output layer and the reservoir. In this method, the network output is fed back to the reservoir through an external feedback loop to suppress the chaos during training.
Such a feedback loop makes ESN training more difficult because changes in readout weights influence the reservoir dynamics, such that further modifications of readout weights are needed to stabilize the internal states of the reservoir.
FORCE learning differs from traditional gradient-descent methods since the output error is small from the beginning, but the number of modifications needed to maintain the output error small is reduced instead.
Besides, the initial activity of spontaneous chaos in the reservoir can make the training more accurate, rapid, and robust \cite{Luko2009Reservoir}.
Although the FORCE learning greatly expands the capabilities of trained RNNs, it does not fully exploit the potentialities of the recurrent connectivity \cite{DePasquale2018full}. Research has been done for extending the FORCE learning, where some recent developments can be referred to \cite{DePasquale2018full, Sussillo2012Transferring, laje2013robust, nicola2017supervised, vandesompele2019populations, Beer2019One, tamura2020two, inoue2020designing}.

The readout training of ESNs can be regarded as a linear regression problem, where the convergence of parameter estimators needs a strict condition termed persistency of excitation (PE) \cite{2011Adaptive}.
Composite learning is an innovative technique to guarantee parameter convergence under a condition of interval excitation (IE) that is strictly weaker than PE, where the rate of parameter convergence can be made arbitrarily fast by increasing an adaptation gain \cite{pan2015composite, pan2017composite, pan2018composite}. The key features of composite learning include dynamic regressor extension and memory data exploitation, which are beneficial to relax the strict requirements on data richness and learning duration for parameter convergence.
Memory regressor extension (MRE) termed in \cite{Gerasimov2020On} can be regarded as a key step in composite learning. Some recent applications of composite learning to real-world control problems can be referred to \cite{guo2018composite, dong2020composite, guo2020composite, guo2021adaptive, huang2019composite, Guo2022b}.
MRE was applied to extend the least-mean-squares (LMS)-based FORCE learning in \cite{RN278Performance}, where both converging speed and estimation smoothness are enhanced compared with the basic LMS-based FORCE learning.
However, the LMS-based FORCE learning has some drawbacks: 1) It is vulnerable to the adverse effects of forgetting mechanisms, which is manifested as the destruction of dynamical objects from previous trials \cite{Beer2019One}; 2) it belongs to a basic delta rule with a time-dependent scalar learning rate rather than a more complex and precise matrix learning rate in recursive least squares (RLS), such that it converges more slowly and cannot predict more complex outputs \cite{Sussillo2009Generating}.

This study presents an RLS-based composite FORCE learning method to train spontaneous chaotic ESNs for chaotic time-series prediction. The design steps are as follows: First, a spontaneous chaotic ESN with an external feedback loop is constructed; second, a new extended regression equation is generated through multiplying the output error equation by the reservoir activation; third, a stable filtering operator is introduced to filter the obtained extended regression equation, and a generalized prediction error is defined; last, the original prediction error and the generalized prediction error are used to update ESN readout weights together.
The proposed method is compared with the basic RLS-based FORCE learning in \cite{Sussillo2009Generating} and the LMS-based composite FORCE learning in \cite{RN278Performance}. Note that the RLS-based MRE and the chaotic time-series prediction are not considered in \cite{RN278Performance}.

This paper is organized as follows: Sec. 2 describes the proposed method, including the network architecture and the training approach; Sec. 3 describes the details of numerical verification; Sec. 4 gives conclusions. In the whole paper, $\mathbb{R}$, $\mathbb{R}^n$, $\mathbb{R}^+$ and $\mathbb{R}^{m\times n}$ denote the spaces of real numbers, real $n$-vectors, positive real numbers, and real $m\times n$-matrices, respectively, $\mathbb{N}^+$ denotes the set of positive natural numbers, $\|\bm x\|$ denotes the Euclidean norm of $\bm x \in \mathbb{R}^n$, $\textit x_i$ denotes the $i$th element of $\bm x$, where $i$, $m$, $n \in \mathbb{N}^+$.

\section{The Proposed Method}
\subsection{Network Architecture}
The network architecture is shown in Fig. 1. We construct an ESN model that only updates synaptic connections from the reservoir to the output neuron. Then, we build an initially spontaneous chaotic reservoir and set up an external feedback loop. The large-scale and sparsely connected reservoir is accomplished by properly setting network parameters $N \in \mathbb{N}^+$, $p \in (0,1)$ and $g \in \mathbb{R}^+$, denoting the number of neurons, a connectivity factor, and a chaotic factor, respectively. The reservoir can be driven by the feedback loop and neuron activations to train readout weights to suppress the chaotic activity of the reservoir and then generate the target output. This study only considers the basic network with $N$ reservoir neurons and one output neuron.

\begin{figure}[!ht]
  \centering
  \includegraphics[width=3in]{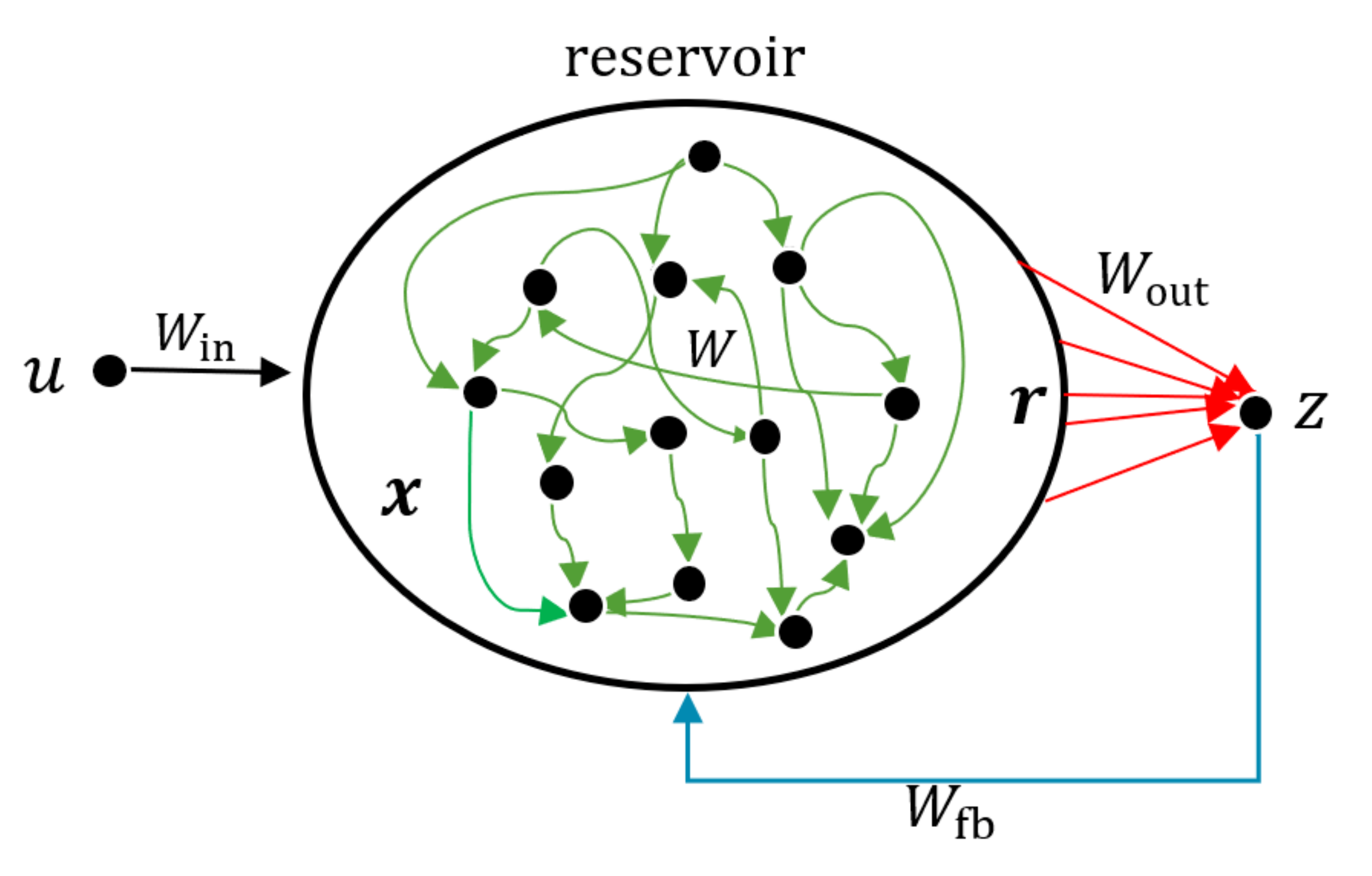}
  \caption{The ESN architecture used in this study, where only red arrows indicate trainable connections.}
  \label{fig:1}
\end{figure}

In Fig. 1, the network input and output are $u(k) \in \mathbb{R}$ and $z(k) \in \mathbb{R}$, respectively, $\bm x(k) \in \mathbb{R}^N$ and $\bm r (k) \in \mathbb{R}^N$ denote the state and activation of reservoir neurons, respectively, $W \in \mathbb{R}^{N \times N}$ is an internal connection weight matrix of the reservoir,
$W_{\rm in} \in \mathbb{R}^{N}$ is a connection weight matrix between the input neuron and the reservoir,
$\hat W_{\rm out} \in \mathbb{R}^{N}$ is an output connection weight matrix,
and $W_{\rm fb}$ $\in \mathbb{R}^{N}$ denotes a feedback connection weight matrix.
The weight matrices $W_{\rm in}$, $W_{\rm fb}$ and $W$ are generated in a pattern of sparse random connectivity and keep fixed after generation, such that only $\hat W_{\rm out}$ needs to be trained. The state $\bm x$ and activation $\bm r$ of the reservoir are updated respectively by \cite{Jaeger2001echostate}
\begin{equation}
\bm x (k) = \phi\big(W_{\rm in} u(k) + W \bm x (k-1) + W_{\rm fb} z(k-1)\big)
\end{equation}
\begin{equation}
\bm r (k) = (1 - \alpha) \bm r (k-1) + \alpha \bm x (k-1)
\end{equation}
where $\phi(\cdot)$ is a nonlinear activation function that works in the element-wise manner, for which we use the tangent function, and $\alpha \in \mathbb{R}^+$ is a leaky rate. The output $z(k)$ is a weighted sum of the neuron activation $\bm r (k)$ as follows:
\begin{equation}
z(k) = \bm r (k)^T \hat W_{\rm out}(k).
\end{equation}

\subsection{RLS-Based FORCE Learning}

FORCE learning is designed to stabilize complex and potentially chaotic dynamics of RNNs by making speedy weight changes under strong feedback. This approach can change the spontaneously chaotic activity of ESNs into a wide range of activity patterns. RLS is a suitable learning algorithm that meets the requirements of FORCE learning, as it can quickly reduce the output error to be a small value and keep it small while searching for optimal readout weights that can maintain a small output error without further modification \cite{Sussillo2009Generating}.

The RLS modification is given as follows \cite{Sussillo2009Generating}:
\begin{equation}
\hat W_{\rm out}(k) = \hat W_{\rm out}(k-1) - e(k)P(k)\bm r (k)
\end{equation}
where $e(k) \in \mathbb{R}$ represents the error between the actual output $z(k)$ and the target output $f(k) \in \mathbb{R}$:
\begin{equation}
e(k) = \bm r (k)^T \hat W_{\rm out}(k-1) - f(k),
\end{equation}
and $P(k) \in \mathbb{R}^{N \times N}$ is a learning rate matrix updated at the same time as $\hat W_{\rm out}$ according to
\begin{equation}
P(k) = P(k-1) - \frac{P(k-1) \bm r (k) \bm r (k)^T  P(k-1)} {1 + \bm r (k)^T P(k)\ \bm r (k)}
\end{equation}
with $P(0) = I/{a}$, in which $I$ denotes the identity matrix, and $a \in \mathbb{R}^+$ is a learning parameter whose value should be chosen according to the target output $f(k)$, subject to a constraint $a << N$. If $a$ is smaller, learning is faster but sometimes can lead to instability; if $a$ is too large, learning may fail.


\subsection{RLS-Based Composite FORCE Learning}

ESNs trained by the FORCE learning can produce a wide variety of complex output patterns, input-output transformations that require memory, and multiple outputs that can be switched by controlling inputs \cite{Sussillo2009Generating}. However, the FORCE learning does not fully exploit the potentialities of the recurrent connectivity because the degree and form of the modifications are restricted \cite{DePasquale2018full}.
It has been found that networks trained by FORCE learning to perform complex problems need more neurons to match the performance of networks trained by gradient-based methods \cite{DePasquale2018full, triefenbach2010phoneme}.

To solve the above problem, we propose an RLS-based composite FORCE learning method, in which the composite learning is applied to improve learning speed, stability, and transient performance of FORCE learning. The basic idea of it is to generate a new extended regression equation via a stable filtering operator with memory.
Multiplying (5) by $\bm r$, one gets an extended regression equation as follows:
\begin{equation}
\bm r (k) e(k) = \bm r (k) \bm r (k)^T \hat W_{\rm out}(k-1) - \bm r (k) f(k).
\end{equation}
Applying a stable filter $L(z) := \frac{\lambda}{1-(1-\lambda)z^{-1}}$, one gets
\begin{equation}
E(k) = \Omega(k)\ \hat W_{\rm out}(k-1) - Y(k)
\end{equation}
with $E(k) := L\{\bm r (k) e(k)\}$, $Y(k) := L\{\bm r (k)f(k)\}$, and $\Omega(k)$ $:=$ $L\{\bm r (k)\bm r (k)^T\}$, where $\lambda \in \mathbb R^+$ is a filtering constant, and $z$ is a Z-transform operator. The RLS-based composite FORCE learning for the update of $\hat W_{\rm out}$ is given by
\begin{equation}
\hat W_{\rm out}(k) = \hat W_{\rm out}(k-1) - P(k) ( e(k) \bm r (k) - \beta E(k))
\end{equation}
where $\beta \in \mathbb{R}^+$ is a learning rate.
The major distinguishing feature of the above method is to make use of the past history of $\bm r (t)\bm r (t)^T$ provided by the filtering operator $L\{\cdot\}$ to improve the learning performance.

\begin{figure}[!t]
	\subfigure[]{
		\includegraphics[width=3.25in]{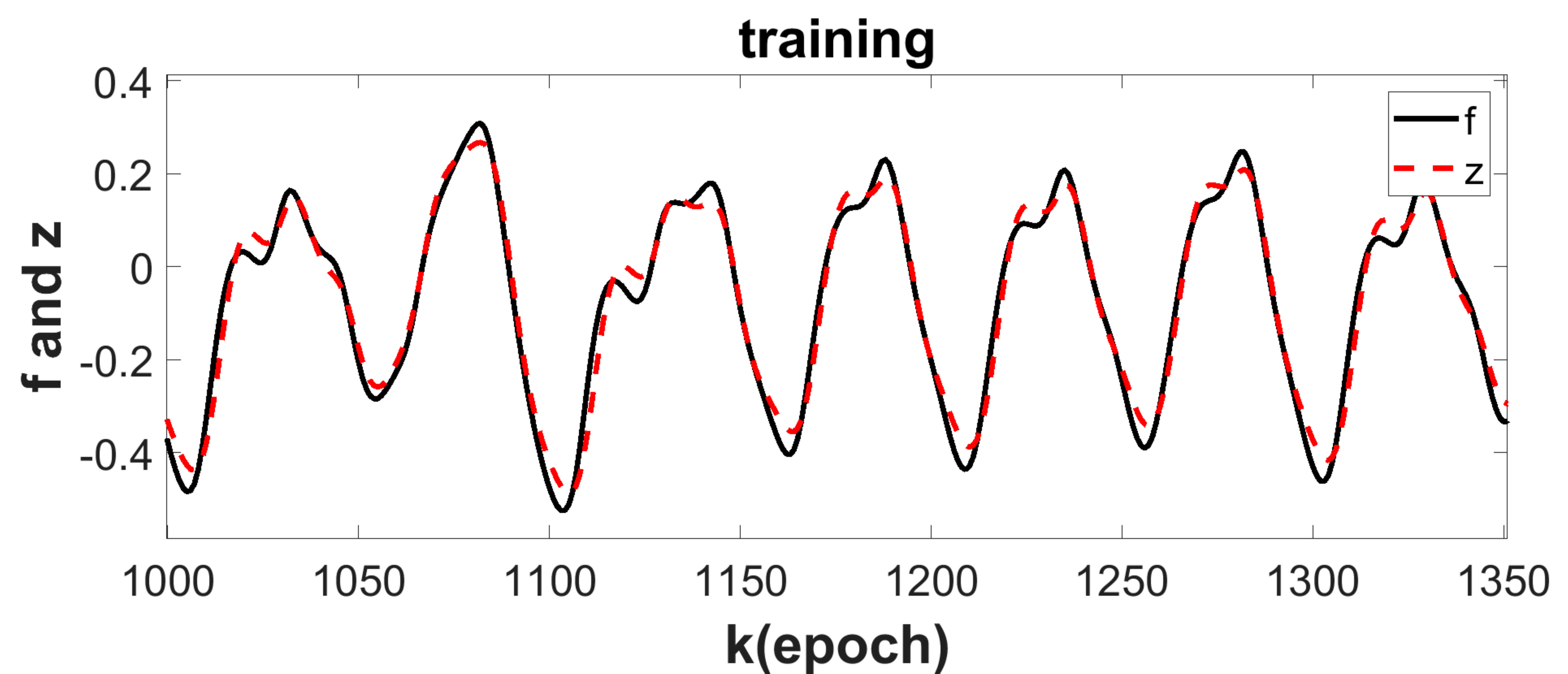}
		\label{training}}\\
	\subfigure[]{
		\includegraphics[width=3.25in]{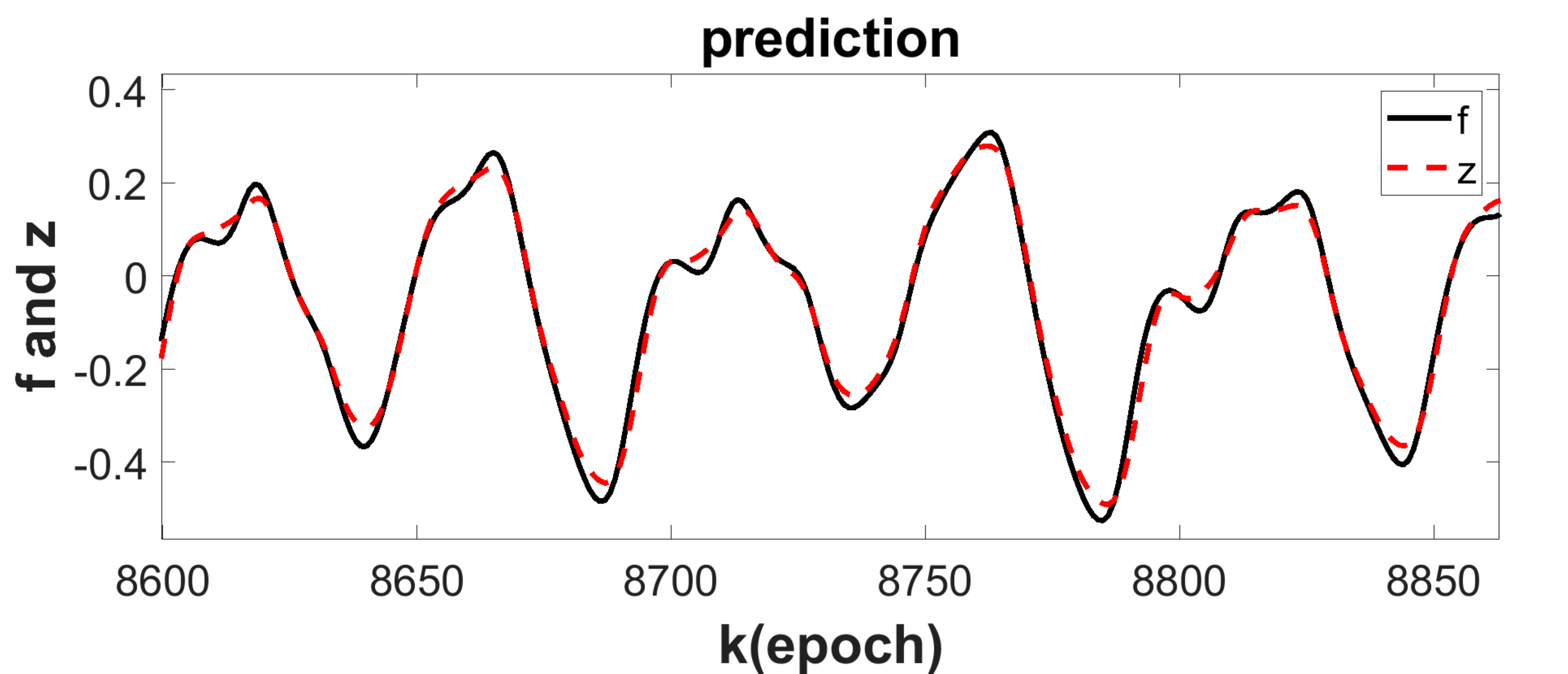}
		\label{test}}\\
	\subfigure[]{
		\includegraphics[width=3.25in]{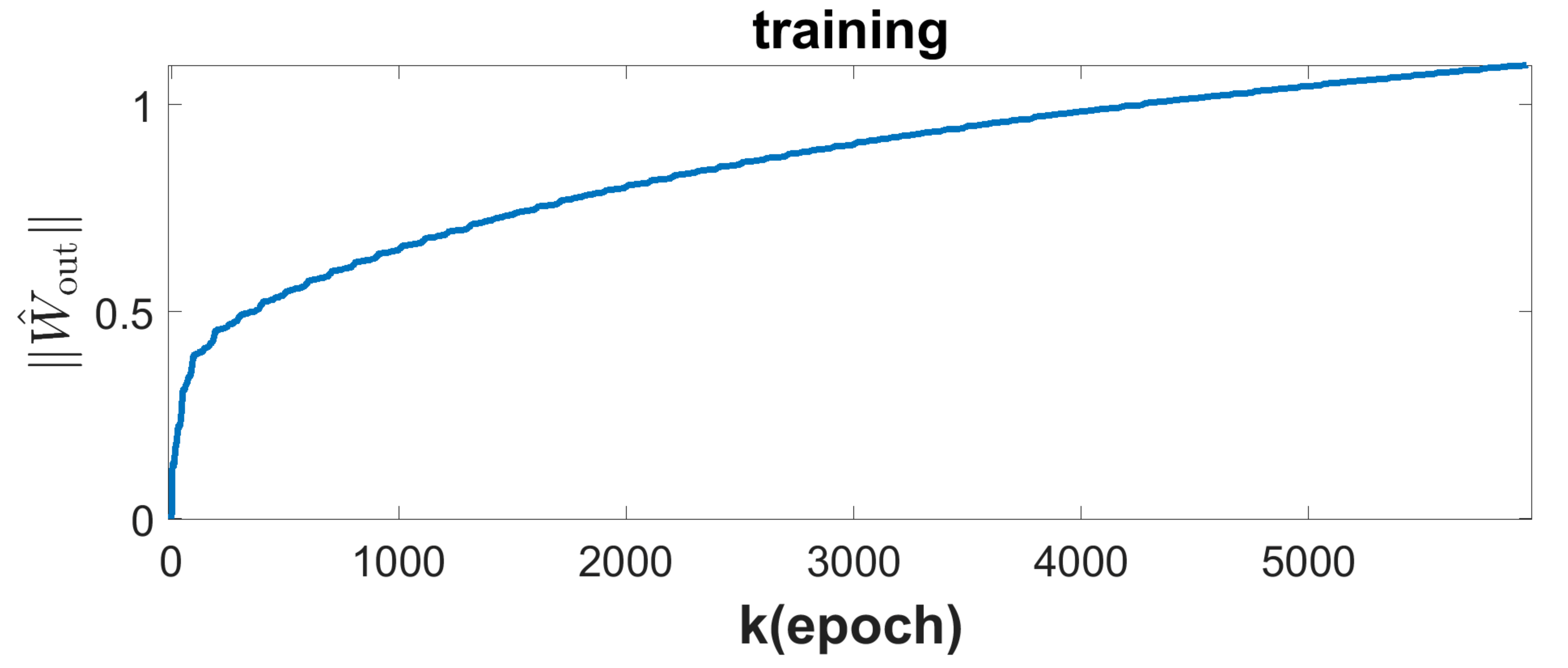}
		\label{3}}
	\caption{Chaotic time-series prediction by the ESN with the basic RLS-based FORCE learning. (a) Training performance, where only a period of the total time is displayed for a clear illustration. (b) Prediction performance, where only a period of the total time is displayed for a clear illustration. (c) The norm of the output weight $\hat W_{\rm out}$.}
	\label{fig:2}
\end{figure}

\begin{figure}[!t]
    \centering
	\subfigure[]{
		\centering\includegraphics[width=3.25in]{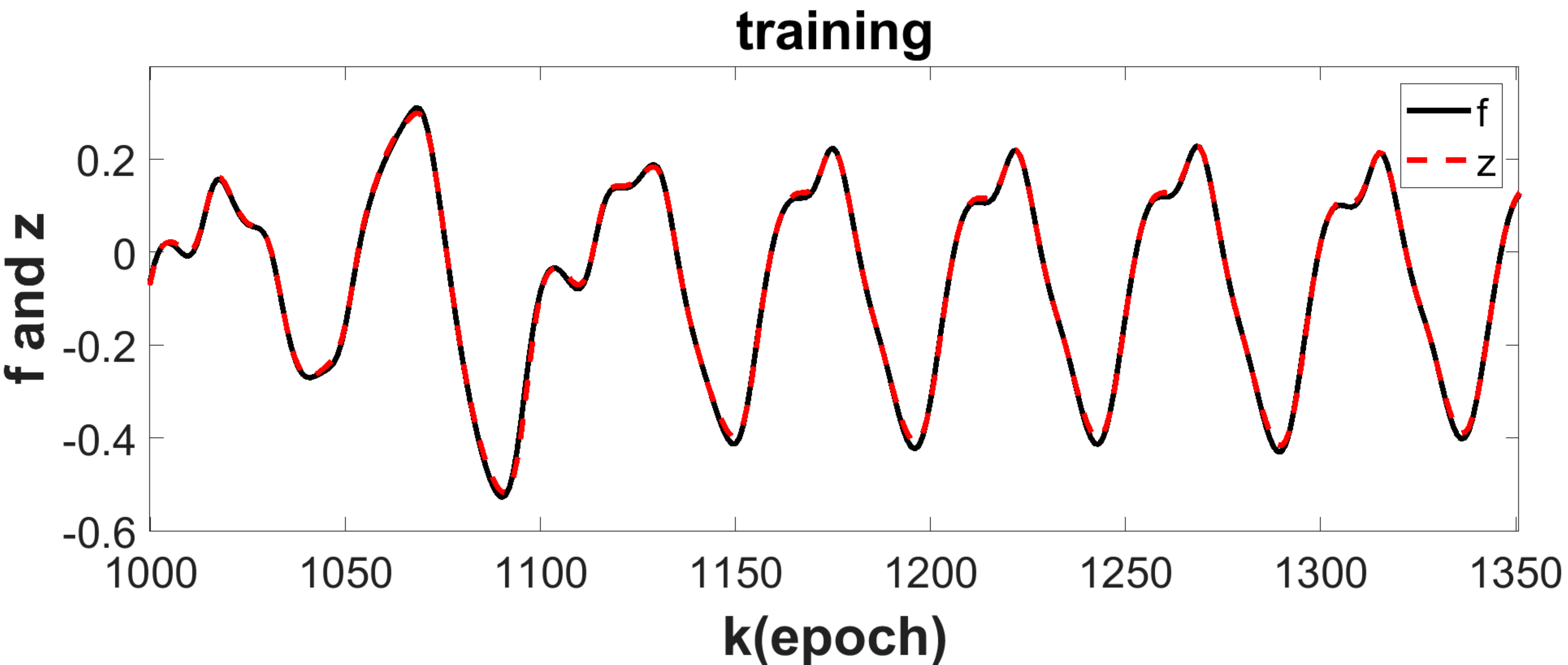}
		\label{training1}}\\
	\centering
	\subfigure[]{
		\centering\includegraphics[width=3.25in]{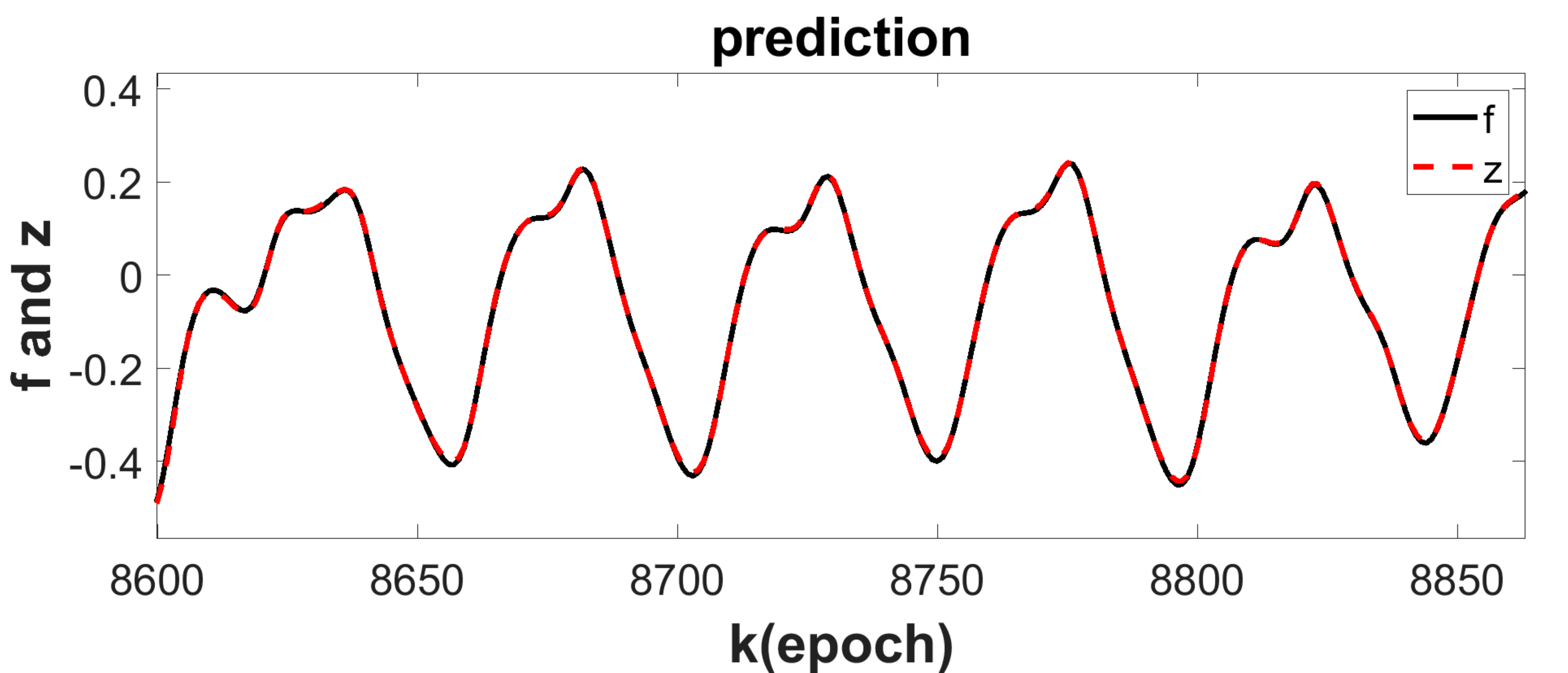}
		\label{test1}}\\
 	\centering
	\subfigure[]{
		\centering\includegraphics[width=3.25in]{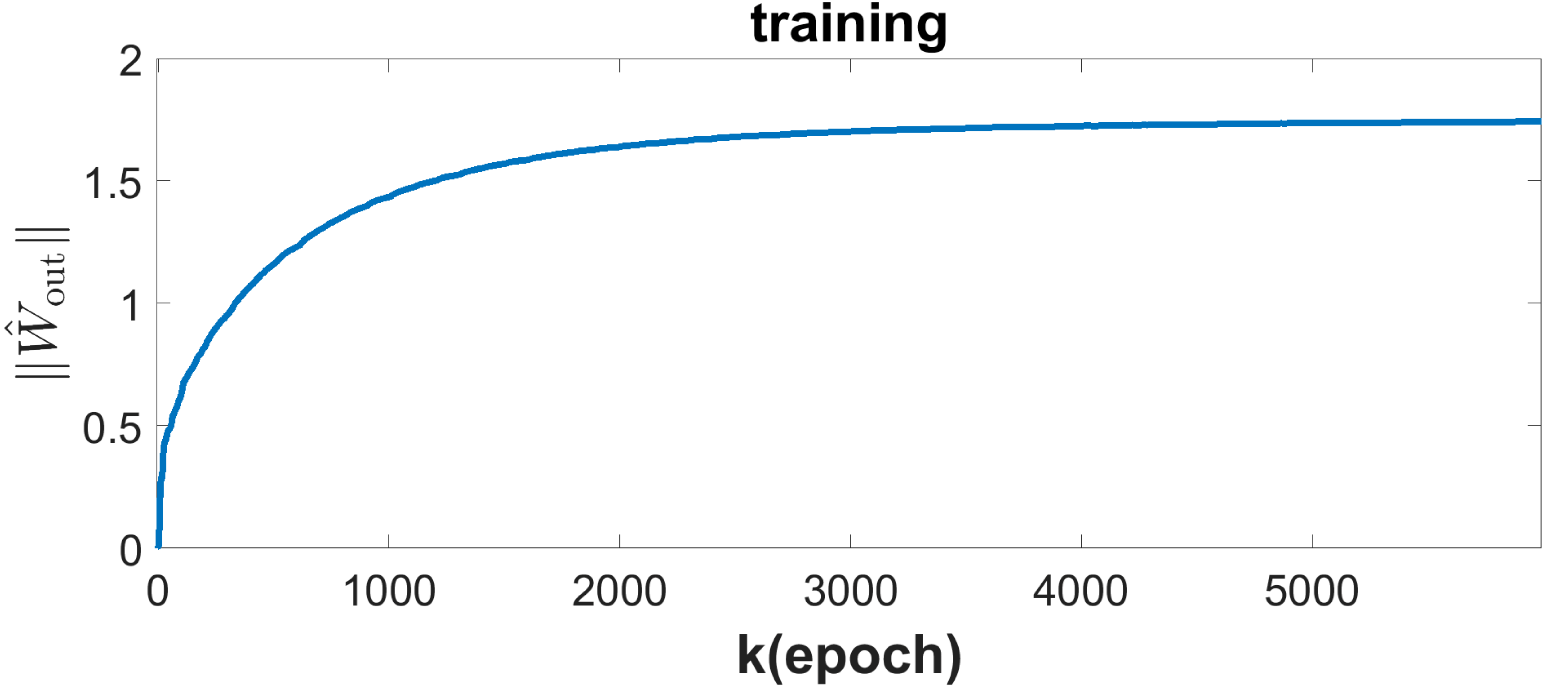}
		\label{ewout2}}
	
	\caption{Chaotic time-series prediction by the ESN with the proposed RLS-based composite FORCE learning. (a) Training performance, where only a period of the total time is displayed for a clear illustration. (b) Prediction performance, where only a period of the total time is displayed for a clear illustration. (c) The norm of the output weight $\hat W_{\rm out}$.}
	\label{fig:3}
\end{figure}

\section{Simulation Studies}
\subsection{Numerical Setup}

We take the task of predicting chaotic time series, which is to train the output weight $\hat{W}_{\rm out}$ of an ESN so that the network output $z(k)$ can keep learning the target dynamic system using chaotic time series generated by itself, and finally the ESN can keep generating target time series on its own after training. For comparisons, the basic RLS-based FORCE learning and the LMS-based composite FORCE learning are used for the same task as the proposed method.

The construction of a chaotic ESN includes the following steps: Firstly, the internal connection weight of reservoir $W$ is generated with a connectivity $p$, where nonzero elements are sampled from a uniform distribution in $[-0.5, 0.5]$, such that the reservoir decomposes into many loosely coupled subsystems, establishing a richly structured reservoir of excitable dynamics;
secondly, the weight $W$ is scaled by the chaotic factor $g$ to make the reservoir chaotic;
thirdly, both the input weight $W_{\rm in}$ and the feedback weight $W_{\rm fb}$ are drawn from a uniform distribution between $[-1,1]$;
finally, the output weight $\hat W_{\rm out}$ is initialized to $\bm0$.


We start simulations in MATLAB platform after the chaotic ESN is constructed according to the above steps. The training time and step size are set as 600s and 0.1s, respectively. When the training epoch reaches a prescribed number (here 6000 epochs), the online learning will be turned off, and the ESN generates predicted series on its own. During simulations, we focus on modeling and prediction performances of the ESN, the convergence of the output weight norm $\|\hat W_{\rm out}\|$, and the evolution of the output weight $\hat W_{\rm out}$. We also calculate the mean squared error (MSE) to make comparison results clearer.
The target chaotic system that generates input-output training sequence is the Mackey-Glass system (MGS), a benchmark system for time series prediction studies. The MGS generates a subtly irregular time series by \cite{Jaeger2004Harnessing}
\begin{align*}\label{MGS}
f(k+1) &=  f(k) + \frac{1}{10} \left[\frac{0.2 f(k-\tau)}{1+f(k-\tau)^{10}} -0.1 f(k)\right]
\end{align*}
where $\tau \in \mathbb R^+$ is a time constant set to 17 in our simulations, and $f(0)$ is initialized to 1.2. The network input sequence $u(k)$ in (1) is set as $f(k)$ generated by the above MGS.

\subsection{Numerical Results}

\begin{figure}[!t]
	\subfigure[]{
		\centering\includegraphics[width=3.25in]{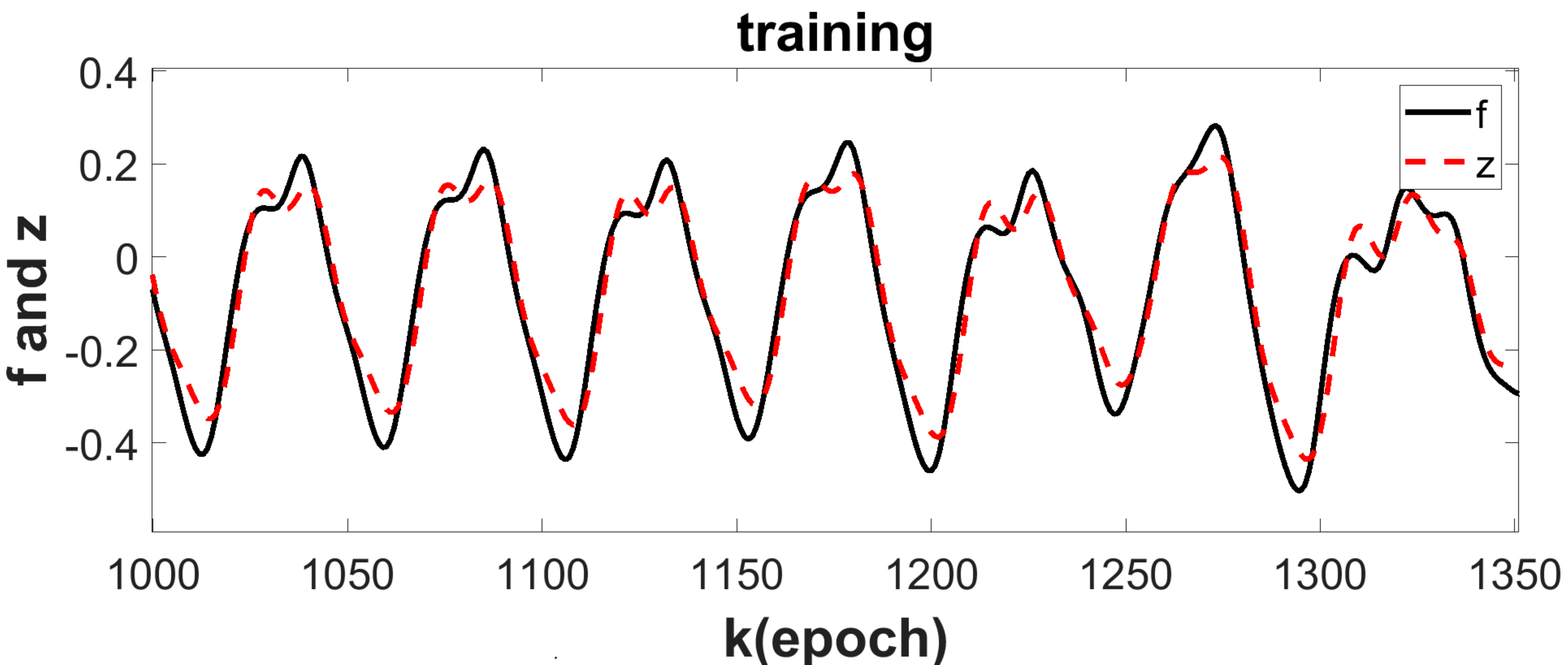}
		\label{training3}}\\
	\subfigure[]{
		\centering\includegraphics[width=3.25in]{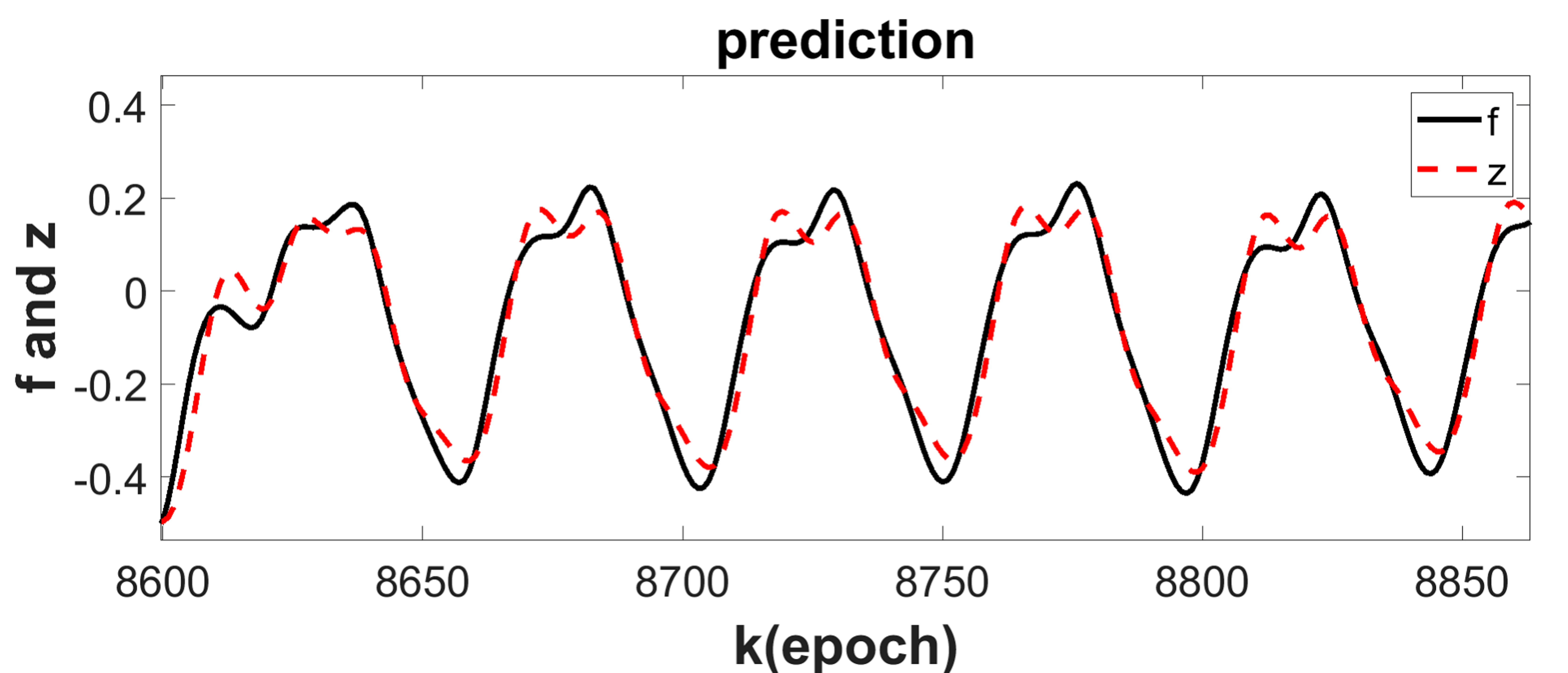}
		\label{test3}}\\
 	\subfigure[]{
		\centering\includegraphics[width=3.25in]{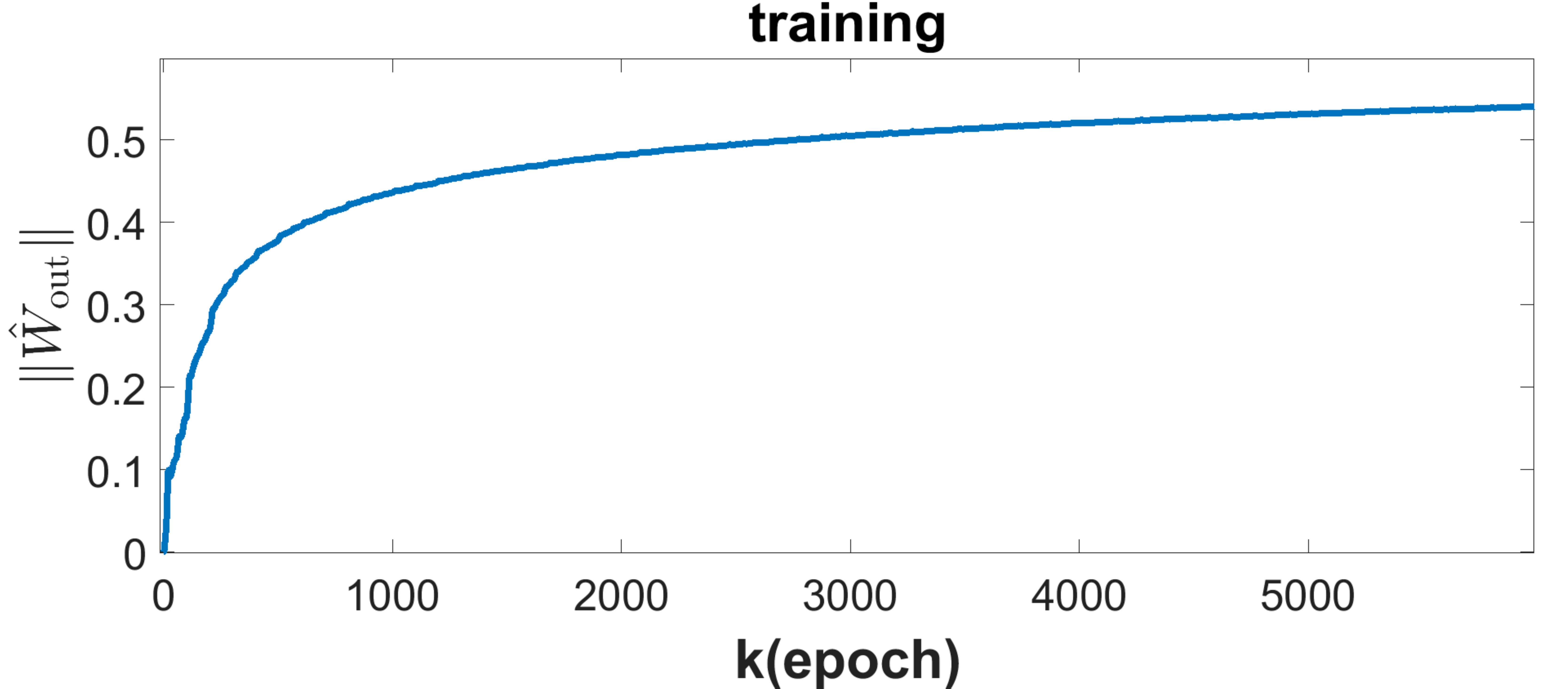}
		\label{5}}
	\caption{Chaotic time-series prediction by the ESN with the LMS-based composite FORCE learning. (a) Training performance, where only a period of the total time is displayed for a clear illustration. (b) Prediction performance, where only a period of the total time is displayed for a clear illustration. (c) The norm of the output weight $\hat W_{\rm out}$.}
	\label{fig:5}
\end{figure}

\begin{figure}[!t]
	\subfigure[]{
		\includegraphics[width=3.25in]{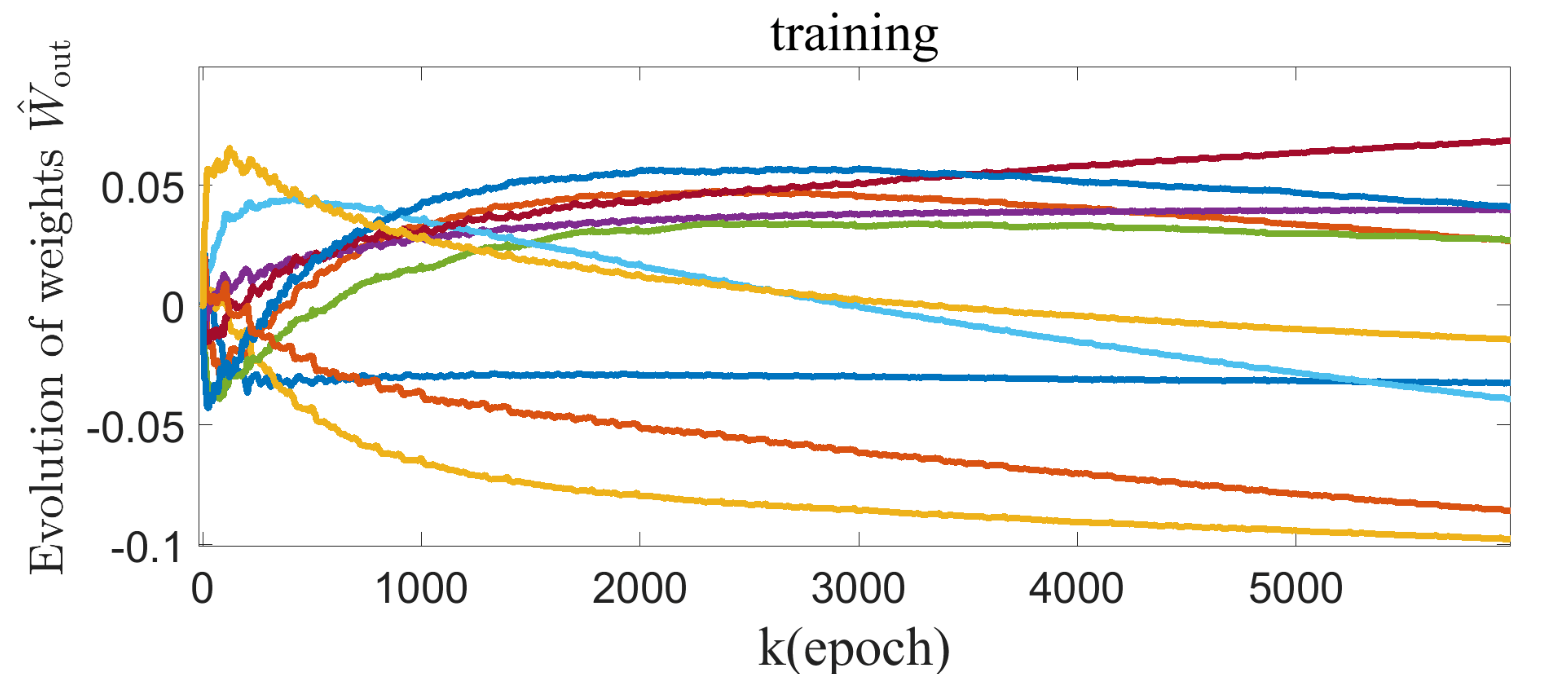}
		\label{Evolution1}}\\
	\subfigure[]{
		\includegraphics[width=3.25in]{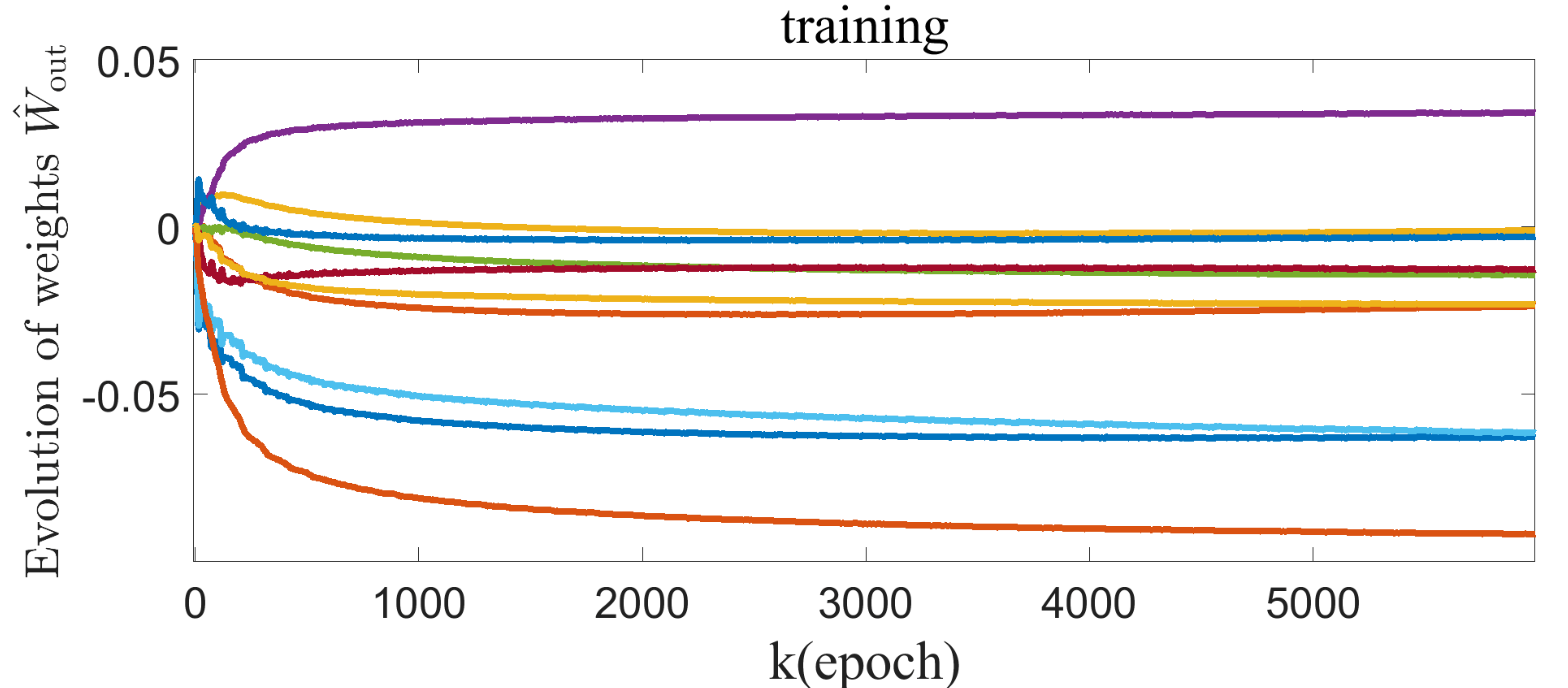}
		\label{Evolution2}}\\
	\subfigure[]{
		\includegraphics[width=3.25in]{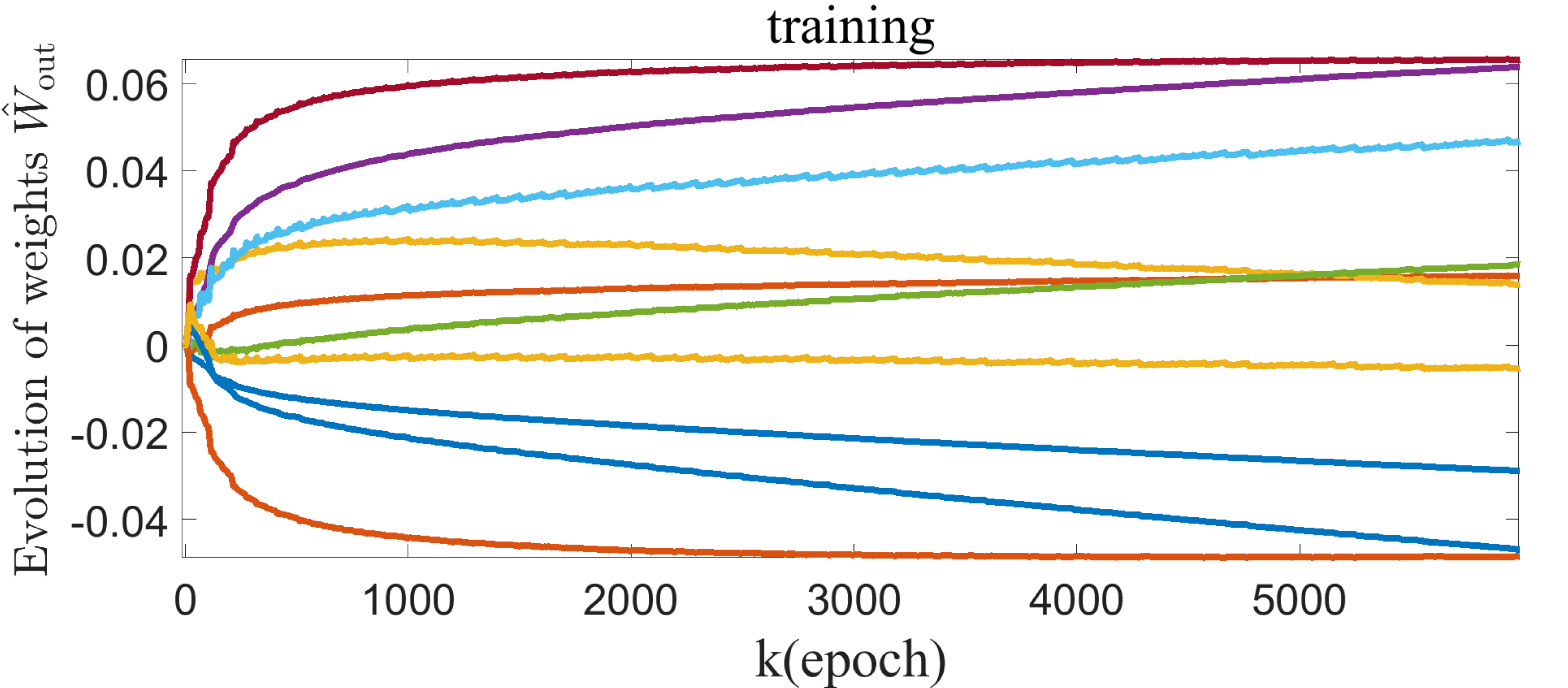}
		\label{Evolution3}}
	\caption{Evolving trajectories of 10 randomly selected elements in the output weight $\hat W_{\rm out}$ under the three learning methods. (a) By the basic RLS-based FORCE learning. (b) By the proposed RLS-based composite FORCE learning. (c) By the LMS-based composite FORCE learning.}
	\label{fig:4}
\end{figure}

The baselines of the proposed method are the basic RLS-based FORCE learning in \cite{Sussillo2009Generating} and the LMS-based composite FORCE learning in \cite{RN278Performance}, where all the shared parameters in the three learning methods keep identical. For the proposed method, set $N=100$, $p=0.1$, $a=1$, $g=2.5$, $\alpha=0.1$, $\beta$ $=$ 3, and $\lambda=0.5$; for the learning method in \cite{Sussillo2009Generating}, set $\beta=0$ and keep the other parameter values the same as the proposed method; for the learning method in \cite{RN278Performance}, we use the same parameter values as the proposed method.

Simulation results of the basic RLS-based FORCE learning are illustrated in Fig. 2. The network output $z(k)$ gradually matches the target output $f(k)$ when the training starts [see Fig. 2(a)]. After the training, the learned ESN can basically reproduce the target system with low prediction accuracy [see Fig. 2(b)]. The output weight norm $\|\hat W_{\rm out}\|$ does not converge to a particular constant after learning [see Fig. 2(c)].

Simulation results of the proposed RLS-based composite FORCE learning are shown in Fig. 3. The network output $z(k)$ rapidly matches the target output $f(k)$ when the training starts [see Fig. 3(a)]. After the training, the learned ESN keeps precisely generating the target output on its own [see Fig. 3(b)]. The output weight norm $\|\hat W_{\rm out}\|$ converges to a constant after only about 2000 epochs [see Fig. 3(c)], which implies that our method is much more time-efficient than the basic RLS-based FORCE learning.
Actually, the basic RLS-based FORCE learning needs 5x more neurons to achieve an effect comparable to the proposed method.

The LMS-based composite FORCE learning in \cite{RN278Performance} improves the modeling performance of ESNs compared with the LMS-based FORCE learning in a dynamic modeling problem with superimposed sinusoidal signals. The method of  \cite{RN278Performance} is applied to the chaotic prediction task of this study that is more complicated, and simulation results are illustrated in Fig. 4. Following the same simulation setting as the proposed method, the network output $z(k)$ roughly matches the target output $z(k)$ [see Fig. 4(a)]. After the training, the learned ESN roughly reproduces the target system, but it is not very accurate [see Fig. 4(b)]. The output weight norm $\|\hat W_{\rm out}\|$ approximately converges to a certain value, but it still changes at the end of learning [see Fig. 4(c)].

The evolution of randomly selected 10 output weight elements in $\hat W_{\rm out}$ for the three methods is shown in Fig. 5. It is clear that the proposed method significantly improves the convergence speed of the output weight $\hat{W}_{\rm out}$.
In addition, the MSE of the three methods shown in Table~\ref{tab1} also implies that the proposed method effectively improves the modeling accuracy of FORCE learning.

\begin{table}[!t]
  \centering
  \caption{MSE Comparison of Three FORCE leaning methods}
  \label{tab1}
  \begin{tabular}{c|c|c}
    \hhline
    \textbf{FORCE learning} & \textbf{Training MSE} & \textbf{Pretection MSE}\\ \hline
    The method of \cite{Sussillo2009Generating} & 0.0223  &0.0136 \\ \hline
    The proposed method & \textbf{0.0093} & \textbf{0.0042} \\ \hline
    The method of \cite{RN278Performance} & 0.0483&0.0366 \\ \hline
    \hhline
  \end{tabular}
\end{table}

\begin{figure}[!t]
	\subfigure[]{
		\centering\includegraphics[width=3.25in]{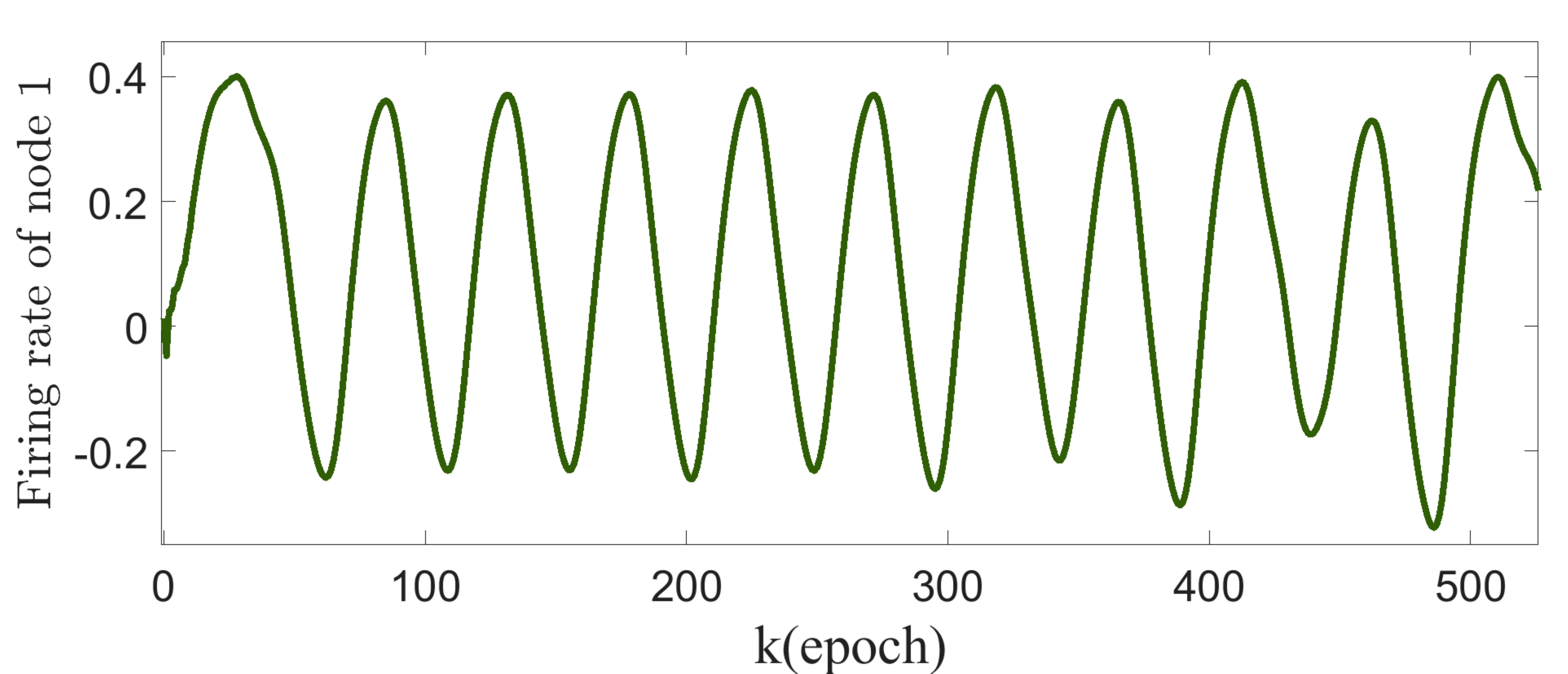}
		\label{NODE1}}\\
	\subfigure[]{
		\centering\includegraphics[width=3.25in]{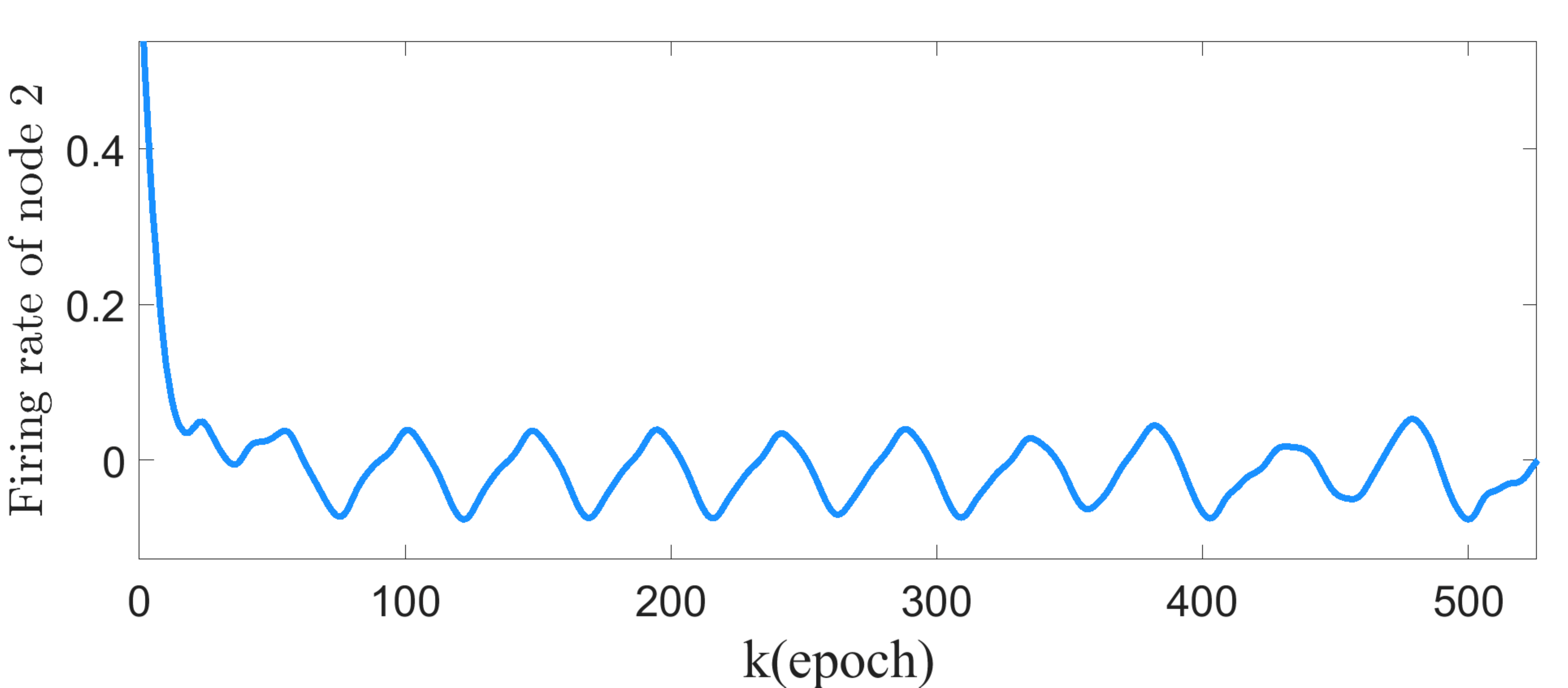}
		\label{NODE2}}\\
 	\subfigure[]{
		\centering\includegraphics[width=3.25in]{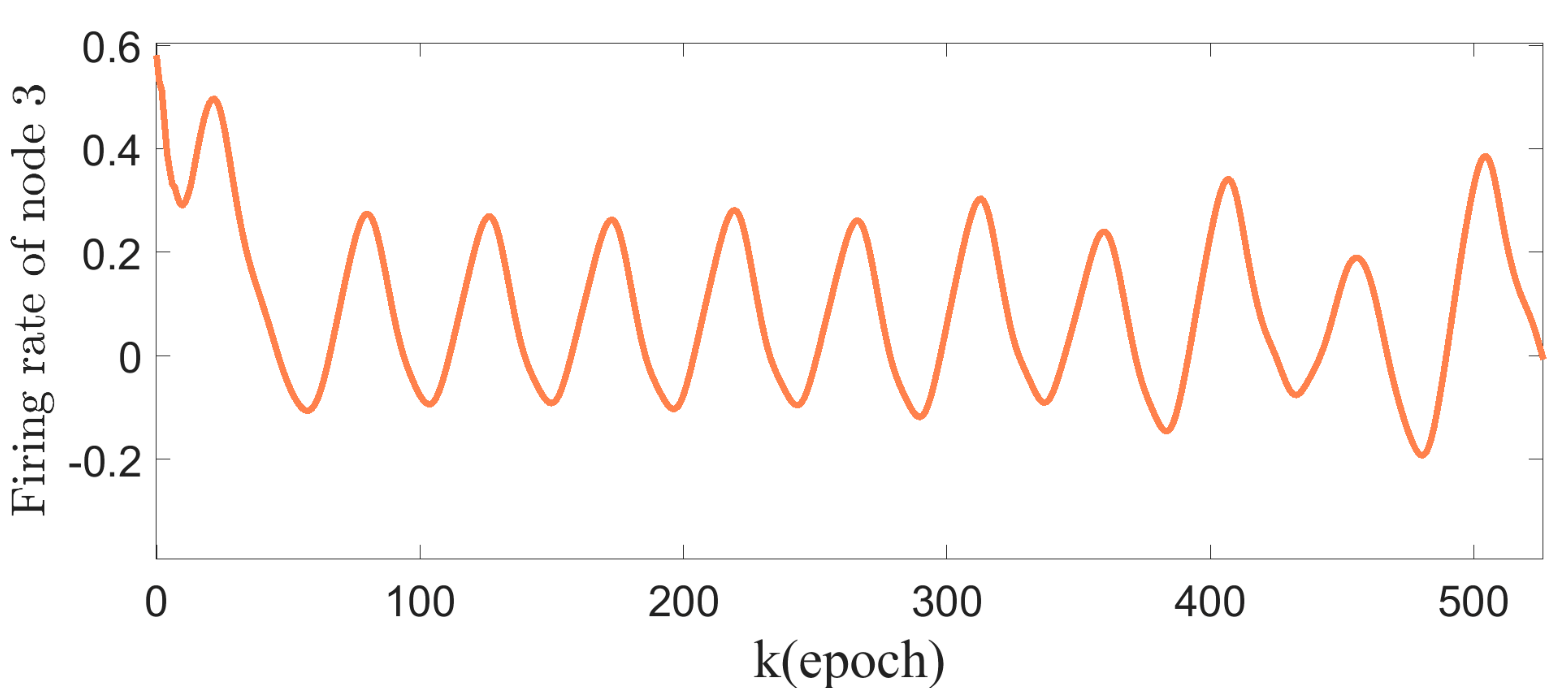}
		\label{NODE3}}
	\caption{The internal dynamics of the reservoir nodes by the proposed FORCE learning, where only 3 randomly selected internal nodes with the first 500 epochs of the total 6000 epochs are displayed for a clear illustration. (a) Evolution of node 1. (b) Evolution of node 2. (c) Evolution of node 3.}
	\label{fig:6}
\end{figure}

The internal dynamics of randomly selected three reservoir nodes for the proposed FORCE learning are illustrated in Fig. 6. The firing rates of the three sampled nodes from the reservoir imply that the network activity is not periodic. This phenomenon reflects the power of the feedback loop that can drive the reservoir to an appropriate dynamic space that matches the requirement of a certain task.

\section{Conclusions}\label{conclusion}

In this paper, an RLS-based composite FORCE learning method has been proposed for chaotic time-series prediction, where the composite learning is applied to enhance learning and prediction performances of FORCE learning. Simulation results on a benchmark example have verified the effectiveness and superiority of the proposed method over the basic RLS-based FORCE learning and the LMS-based composite FORCE learning. More specifically, the proposed method greatly speeds up parameter convergence and improves learning stability on predicting chaotic time series generated by the MGS. Meanwhile, the proposed method contributes to reducing the scale of ESNs. The application of the proposed method to more complicated dynamic modeling problems would be investigated in our further work.

\bibliographystyle{IEEEtran}
\balance
\bibliography{main}

\end{document}